# An Algorithm Based on Empirical Methods, for Text-to-Tuneful-Speech Synthesis of Sanskrit Verse

Rama N.[†] and Meenakshi Lakshmanan[††]

[†]Department of Computer Science, Presidency College, Chennai, India

[††]Department of Computer Science, Meenakshi College for Women, Chennai, India
and
Research Scholar, Mother Teresa Women's University, Kodaikanal, India

**Summary**

The rendering of Sanskrit poetry from text to speech is a problem that has not been solved before. One reason may be the complications in the language itself. We present unique algorithms based on extensive empirical analysis, to synthesize speech from a given text input of Sanskrit verses. Using a pre-recorded audio units database which is itself tremendously reduced in size compared to the colossal size that would otherwise be required, the algorithms work on producing the best possible, tunefully rendered chanting of the given verse. His would enable the visually impaired and those with reading disabilities to easily access the contents of Sanskrit verses otherwise available only in writing.

*Key words:*

*Sanskrit, verse, text-to-speech, musical tones, speech synthesis, sandhi, metre.*

## 1. Introduction

Speech synthesis systems have proved to be extremely useful in improving the lives of the visually impaired and those with reading disabilities across the globe. However, such systems that cater to western language are not applicable in the Indian context, because of the huge difference in the structure and pronunciation schemes of Indian languages. Work has been done to bring Indian vernaculars to the people through speech synthesis [3, 16, 18], but there is a dearth of such work in the context of Sanskrit.

Even a cursory glance at randomly chosen works in the Sanskrit literature would reveal that poetry hugely dominates the literature. The volume of the extant literature is vast and the contents profound, with topics ranging from grammar to spirituality, from medicine to geography. Listening to verses being chanted and committing them to memory has been a traditional practice. Similar is the case with chanting them tunefully.

Obviously, rather good familiarity with the Sanskrit script is required to read the verses, and that too continuously, with reasonable speed and with a tune. Thus, the visually impaired and those with reading disabilities would find themselves at a serious disadvantage, as would those who do not know how to read Sanskrit but would like to know or memorize verses. Further, in today's fast-moving world in which time is at a premium, a piece of software that reads out any desired Sanskrit poetical text would be welcome.

We propose a comprehensive method based on empirical analysis, to convert Sanskrit poetical text to speech. This method is new, effective and produces output that is tuneful.

## 2. The Problem

The most important qualities of a speech synthesis system are naturalness and intelligibility. Naturalness describes how closely the output sounds like human speech, while intelligibility is the ease with which the output is understood [7]. The most popular and simplest method of speech synthesizing is the concatenative method. Formant synthesis, the other major speech synthesis method, would inevitably compromise on naturalness of the voice output [5, 9, 12, 14]. We deal with only the concatenative method in this work.

Sanskrit is a highly phonetic language, which adheres completely to the "what you see is what you hear" rule. Further, it is highly structured with stringent rules in its phonemic and morphological levels, but lends itself to extreme versatile in the higher syntactic, semantic and pragmatic levels. As such, the limit for framing new compound words in Sanskrit is only the poet's imagination and linguistic skill. This fact coupled with the complications posed by *sandhi*-s and case-inflectional forms or *vibhakti*-s, ensures that the standard method of text-to-speech synthesis, viz. creating a voice database of words in the dictionary and concatenating these stored audio files while parsing the verse, is well nigh impossible to apply in the case of Sanskrit, in spite of the highly phonetic nature of the language. Secondly, creating just the pronounced individual letters as audio files and then concatenating them while parsing the verse, would render a rather poorly pronounced verse, and in fact an incorrectly pronounced one.

Consider the sample verse,

> vande gurūṇāṁ caraṇāravinde
> sandarśitassvātmasukhāvabodhe |
> janasya ye jāṅgalikāyamāne
> saṁsārahālāhalamohaśāntyai ||

The large compound word,

> sandarśitassvātmasukhāvabodhe

is actually *sandarśitaḥ* + *svātmasukhāvabodhe* to which a *sandhi* rule has been applied. There is no way one could have stored a priori, this entire compound word created on the fly by the poet. Similarly, the word *janasya* comes from the word *janaḥ* which would be found in a dictionary, unlike *janasya*. The reason is that the sixth out of eight case-endings has been applied to the root word *janaḥ* meaning "people", resulting in *janasya* meaning "of people". There are 24 such case-inflectional forms in total for every noun, and nine or eighteen for verbs in each of six tenses and four moods. Considering that the count of nouns and verbs is in the thousands, the storing and retrieval of audio snippets of all such case-inflectional forms would be prohibitive in terms of space and time for creation and retrieval.

Pronouncing the verse letter by letter would give "*v + a + n + d + e*", etc., which is incorrect pronunciation. Even if the unit of pronunciation be considered as a consonant with its succeeding vowel alone, it is insufficient, for "*va*" would be pronounced correctly, but there would be a problem again with streaming the "*n*" separately, for it would result in weird pronunciation.

It is thus clear that none of the methods of speech synthesis outlined above would be effective in the case of Sanskrit text-to-speech processing. We present algorithms to make the output intelligible and constituting a correct reading of the verse with pauses as per the caesura data and also tunefully.

## 3. The Precursor to this Work

Euphonic conjunctions or *sandhi*-s in Sanskrit are points between adjacent words or sub-words at which letters coalesce and transform. The application of *sandhi* is compulsory in Sanskrit verse, though the rules are not as stringent in the prose. A novel computational approach to *sandhi* processing based on building *sandhi*-s rather than splitting them, was developed by the authors [10]. This was done in accordance with the grammatical rules laid down by the ancient Sanskrit grammarian-genius Pāṇini in his magnum opus, the *Aṣṭādhyāyī* and forms a comprehensive *sandhi*-building engine.

An example of *sandhi* is: *namaḥ* + *śivāya* = *namaśśivāya*. The *visarga* letter (*ḥ*) gets transformed because of the presence of the letter *ś* after it, into *ś*. This is an example of consecutive application of a *visarga sandhi* rule and the *ścutva sandhi* rule [10]. Though the original words *namaḥ* and *śivāya* are independent words and are per se correct as they are, the rules of verse demand that the *sandhi* at their junction be applied and the transformation done as shown. This becomes relevant in the context of speech synthesis, because after the application of the *sandhi* rule, the two words become one compound word and, as a result of the doubling of the letter *ś*, is read with a stress on the letter.

Secondly, verses in Sanskrit are classified into metres according to the number and type of syllables in the four quarters of the verse. Algorithms to efficiently parse and classify verses into more than 700 metres and to gather information about the caesura or points in the verse where a pause must be introduced while reading the verse, were developed by the authors [11]. Verses of different metres are read in different tempos, with pauses at different caesura and with different tunes. Hence the information provided by the metrical classification algorithm already developed by the authors to handle input verses in both Sanskrit Unicode and as E-text in the Latin character set, is an important input for this work.

## 4. Text Pre-processing

### 4.1 Unicode Representation of Sanskrit Text

The Unicode (UTF-8) standard is what has been adopted universally for the purpose of encoding Indian language texts into digital format. The Unicode Consortium has assigned the Unicode hexadecimal range 0900 - 097F for Sanskrit characters.

All characters including the diacritical characters used to represent Sanskrit letters in E-texts are found dispersed across the Basic Latin (0000-007F), Latin-1 Supplement (0080-00FF), Latin Extended-A (0100-017F) and Latin Extended Additional (1E00 – 1EFF) Unicode ranges.

The Latin character set has been employed in this paper to represent Sanskrit letters as E-text.

The text given in the form of E-text using the Unicode Latin character set, is taken as input for processing. Unicode Sanskrit font is also accepted as input, but is converted to the Latin character form before processing begins, as already presented by the authors in [11].

### 4.2 *Sandhi* Correction

The following *sandhi* rules are specifically relevant because the transformation wrought by them have a bearing on the pronunciation of the word.

1. For the letter combination "*hn*", such as in the word "*vahni*", the normal pronunciation is actually "*nh*", i.e. as "*vanhi*". Thus, when the combination "*hn*" is encountered in a word, it is replaced by "*nh*".
2. *Sandhi* rules with respect to the *anusvāra* are applied. For example, the word "*saṁnyāsa*", split as "*saṁ+nyāsa*" is normally not pronounced this way. Instead, it is pronounced as "*sannyāsa*", the transformation being governed by a *sandhi* category called the *parasavarna sandhi*.
3. *Sandhi* rules involving the transformation of the *visarga*. The example "*namaḥ + śivāya*" discussed earlier in Section 3 is a typical one.
4. *Sandhi* rules for the *jihvāmūliya* (the aspirate sound produced near the base of the tongue while pronouncing the *visarga* that is followed by '*k*' or '*kh*') and the *upadhmānīya* (the sound of 'f' while pronouncing the *visarga* that is followed by '*p*' or '*ph*') have to be applied for correct pronunciation. At the text pre-processing stage, such *visarga*-s are replaced appropriately by 'z' for the *jihvāmūlīya* and by 'f' for the *upadhmānīya*.

The algorithm in [10] serves to effect these corrections on the given verse.

### 4.3 Identifying the Syllabic Units of the Verse

The text being processed has to be divided into single-syllabic units at the pre-processing stage, in such a way that each of the units is pronounceable. This is riddled with problems and cannot be handled like European languages [1, 2, 4, 6, 8], as already discussed in Section 2 above. We refer to such pronounceable units as just 'units'.

A unit can have a maximum of three components:

1. Vowel component
2. Pre-vowel component
3. Post-vowel component

The vowel component is central and indispensable to the unit. A unit would have the vowel component and optionally one or more of the other two components. Further, the pre-vowel and post-vowel components may consist of one or more characters.

For our purposes, we use the categorization of the Sanskrit alphabet given in Table 1.

Table 1: The Sanskrit alphabet categorized

| # | Category | Letters |
|---|---|---|
| 1 | Vowels | a, ā, i, ī, u, ū, ṛ, ṝ, ḷ, e, ai, o, au |
| 2 | Short vowels | a, i, u, ṛ, ḷ |
| 3 | Long vowels | ā, ī, ū, ṝ, e, ai, o, au |
| 4 | Consonants | k, kh, g, gh, ṅ |
|   |   | c, ch, j, jh, ñ |
|   |   | ṭ, ṭh, ḍ, ḍh, ṇ |
|   |   | t, th, d, dh, n |
|   |   | p, ph, b, bh, m |
| 5 | Semi-vowels | y, r, l, v |
| 6 | Sibilants | ś, ṣ, s |
| 7 | Aspirate | h |
| 8 | *Anusvāra* | ṁ |
| 9 | *Visarga* | ḥ |

The following empirically determined cases constitute all the possibilities that arise while parsing a verse:

1. We parse the verse starting from the end of the last unit identified, until we encounter the first vowel. As we do so, we include all the letters on the way in the unit.
2. If a consonant is encountered and it happens to be from the first or third columns of the Consonants category shown in Table 1, then we have to examine the following letter to see if it is an '*h*'. If it is, then the two letters together constitute a consonant belonging to the second or fourth rows of the Consonants category. This is important in order to correctly determine the next letter while parsing.
3. If the vowel encountered = '*a*' and is followed by '*i*' or '*u*', then the vowel is really '*ai*' or '*au*' respectively.
4. If a *visarga* or *anusvāra* follows the vowel of the unit, then it is included in the unit and the unit is closed.
5. If the vowel is followed by a consonant that is in turn followed by a vowel, then the unit is closed with the vowel itself. **Eg:** In the word "*gurūṇāṁ*", first '*g*' is taken, and then the vowel '*u*' is encountered. Now since the vowel of the unit is followed by a consonant ('*r*') and then a vowel ('*ū*'), the unit is closed as "*gu*". The next unit will begin with the '*r*'.
6. If the vowel is followed by a consonant and then by a non-vowel, then the consonant is also included in the unit and the unit closes with that. **Eg:** In the word "*vande*", after parsing upto "*va*", it is found that the vowel '*a*' of the unit is followed by a consonant ('*n*') that is in turn followed by a non-vowel ('*d*'). Hence, the unit includes '*n*' also and is closed as "*van*".

7. If the vowel is followed by the letter '*r*', followed by a non-vowel and again a non-vowel (and any number of such non-vowels), then the unit is taken to include the letter '*r*' and the non-vowel following it. **Eg:** In the word "*kārtsnyaṁ*", after "*ka*" is parsed, we encounter '*r*' followed by a non-vowel ('*t*') and again a non-vowel ('*s*'). Thus, the unit is taken as "*kārt*". Indeed, the word is pronounced as *kārt-snyaṁ*.

8. If the vowel is followed by '*r*', then a non-vowel and then a vowel, then the unit is closed with the '*r*'. **Eg:** In the word "*kāryaṁ*", "*kā*" is parsed, and after the following '*r*', we have a non-vowel ('*y*') and then a vowel ('*a*'). Hence the unit is closed as "*kār*". The way the word is pronounced in Sanskrit is *kār-yaṁ*, and hence this is valid.

**Exceptions to Rule 6:**

a. Whenever the vowel is followed by the consonant pairs "*jñ*" or "*kṣ*" then the unit is closed with the vowel itself. **Eg:** In the word "*ajñā*", the pronunciation is *a-jñā* and not *aj-ñā* as Rule 6 would have it.

b. In cases where the consonant pairs "*pr*" or "*br*" or "*kr*" or the consonant '*h*' follow a short vowel, the unit is closed with the vowel itself. **Eg:** The word "*sapriyaḥ*" is to be pronounced *sa-priyaḥ* and not as *sap-riyaḥ* as would be required by Rule 6.

We propose the following parsing algorithm that parses a given verse, handles all the above cases including the exceptions and recognizes the units in the verse.

**Algorithm** SplitVerseIntoUnits

```
//strVerse is a string variable that contains the entire verse.
//strVerse(i) denotes its i-th character.
//strUnit is a string variable storing the unit being
//processed. It is initialized to the empty string.
i = 0;
while i < strVerse.length() do
    while strVerse(i) is not a vowel do //parse till 1st vowel
        append strVerse(i) to strUnit;
        i = i + 1;
    end while
    if k = 'a' and (k₁ = 'i' or k₁ = 'u') then
        append k₁ to strUnit;
        i = i + 1;
    end if
    k = strVerse(i); //the first vowel in the unit
    k₁ = strVerse(i+1);
    if k₁ is a consonant
        if (k₁ = 'k' or k₁ = 'g' or k₁ = 'c' or k₁ = 'j' or k₁ =
        't' or k₁ = 'd' or k₁ = 't' or k₁ = 'd' or k₁ = 'p' or k₁
        = 'b') and strVerse(i+2) = 'h' then
            i = i + 1;
        end if
    end if
    k₂ = strVerse(i+2);
    if k₂ is a consonant
        if (k₂ = 'k' or k₂ = 'g' or k₂ = 'c' or k₂ = 'j' or k₂ =
        't' or k₂ = 'd' or k₂ = 't' or k₂ = 'd' or k₂ = 'p' or k₂
        = 'b') and strVerse(i+3) = 'h' then
            i = i + 1;
        end if
    end if
    k₃ = strVerse(i+3);
    append k to strUnit;
    if k₁ is visarga or k₁ is anusvāra then
        append k₁ to strUnit;
        close strUnit and initialize to empty string to hold
        the next unit;
    else if k₁ is a consonant
        if k₂ is a vowel then
            close strUnit and initialize to empty string to
            hold the next unit;
        else if (k₂='j' and k₃='ñ') or (k₂='k' and k₃='ṣ') then
            close strUnit and initialize to empty string to
            hold the next unit;
        else if (k is a short vowel) and (k₂k₃ = "pr" or k₂k₃
        = "br" or k₂k₃ = "kr" or k₂ = 'h') then
            close strUnit and initialize to empty string to
            hold the next unit;
        else //k₂ not vowel, above exceptions do not hold
            append k₁ to strUnit;
            close strUnit and initialize to empty string to
            hold the next unit;
        end if
    else if k₁ is 'r' and k₂ is not a vowel
        if k₃ is not a vowel then
            append k₁ to strUnit;
            append k₂ to strUnit;
            close strUnit and initialize to empty string to
            hold the next unit;
        else //k₃ is a vowel
            append k₁ to strUnit;
            close strUnit and initialize to empty string to
            hold the next unit;
        end if
    end if
    i = i + 1;
end while
```

end **Algorithm**

## 5. The Audio Units Database

Determining the units to be recorded as audio files and stored in a database is a non-trivial task, as is clear from the discussions in Sections 2 and 4 above. The possible number of such recordable units is huge. There are a total of 34 letters in the consonants, semi-vowel, sibilant and aspirate categories, and 13 vowels. As such, for the units with one consonant and one vowel, we would have possible 34 x 13 = 442 recordable units, which seems feasible. But for units having two consonants, a vowel and a consonant, there are theoretically 34 * 13 * 34 * 34 = 510, 952 recordable units possible! The number of cases with more than two consonants in close succession would obviously be much greater. Thus, we see that there is a combinatorial explosion of the number of required recordable units in this scheme.

This problem was surmounted by extensively analyzing the words present in a comprehensive Sanskrit dictionary [13]. Pronounceable units were gathered through this exercise, and the total number of units for practical use was thus substantially reduced. Further, the glyphs of the Sanskrit 2003 font were also studied and some possible units were eliminated as unpronounceable. This font was particularly chosen because it provides a plethora of composite glyphs as well. In this manner, the number of readable units was significantly reduced to yield a database size of just approximately 2000 recorded unit clips.

The recording of each unit was done in the same vocal pitch. However, the length of intoning of the sound was varied according as whether the unit being intoned has a long (*guru*) vowel or a short (*laghu*) one. All long vowels are *guru* and all short vowels are *laghu*. The exception is that *laghu* vowels become *guru* if they are

1. followed by double consonants (this being optional in the case of *pr, br, kr* and *h*)
2. followed by anusvāra or visarga

All *laghu* units were recorded in one time unit, and all *guru* ones, in two. For example, in the sample verse given in Section 2 above, the word "*vande*" has two units, "*van*" and "*de*", which are respectively of the *laghu* and *guru* kinds. As such, "*van*" was recorded in one time unit (2 seconds) and "*de*" in two time units (4 seconds).

The reason for using 2 and 4 seconds rather than perhaps 1 and 2 seconds, is that a units have different number of letters in them and yet have to be intoned in the same time span. For example, the units *kā* and *kārt* are both *guru* units, and hence must each be intoned in two time units. However, it would clearly take a little longer to pronounce the whole of *kārt* than it would to pronounce *kā*. Hence, the *kā* would have to elongated and recorded. It is to provide for this that the longer time spans of 2 and 4 seconds were fixed and followed during the recording process.

The purpose of assigning the duration of an audio clip based on whether the unit being pronounced is *laghu* or *guru*, is to help make the synthesized chanting of the verse follow a beat. This factor is important in order to achieve near-human reproduction of verse-chanting. The beat to be followed for a verse would vary according to the metre and associated caesura of the verse [11].

## 6. The Musical Component of the Speech Synthesizer

Indian music is similar to its Western counterpart in the context of the theory of notes and octaves. There are a total of 12 notes in each octave, with the notes separated by a few frequencies. The basic notes bear the names *sa, ri* (soft)*, ri, ga* (soft)*, ga, ma, ma* (sharp)*, pa, dha* (soft)*, dha, ni* (soft)*, ni*. These 12 notes repeat themselves in higher and higher levels of frequencies, and thus are octaves formed.

To make the chanting of the verse tuneful, it is necessary to introduce these musical notes. The fundamental idea is that slight changes made to the frequency of the recorded audio unit file, results in a change of the musical note at which the unit is heard when played. The changes have necessarily to be slight, for otherwise it is found that the very texture of the voice changes with vast changes in frequency.

Table 2: Values given for musical notes

| # | Note | Value |
|---|------|-------|
| 1 | sa | -7 |
| 2 | ri (soft) | -6 |
| 3 | ri | -5 |
| 4 | ga (soft) | -4 |
| 5 | ga | -3 |
| 6 | ma | -2 |
| 7 | ma (sharp) | -1 |
| 8 | pa | 0 |
| 9 | dha (soft) | 1 |
| 10 | dha | 2 |
| 11 | ni (soft) | 3 |
| 12 | ni | 4 |

Let $u_1, u_2, \ldots u_n$ be the n units constituting the given verse. Now the two variables associated with each such unit are the syllable intoned and the pitch of the sound. We will assume that the same octave is maintained throughout. Let $p_i$ denote the pitch at which $u_i$ should be intoned in order for the verse to be chanted tunefully. Thus, the final output for the ith unit is a function $f$ of $u_i$ and $p_i$. Thus, the final speech output is not just $\sum_{i=1}^{n} u_i$ but $\sum_{i=1}^{n} f(u_i, p_i)$ where summation here stands for concatenation.

The pitch levels $p_1, p_2, \ldots p_n$ for the n syllables of a verse's quarter, were fixed for the categories of metres enumerated and stored in the database created for the earlier work on metre classification outlined in Section 3 [11]. Here, n may vary from 1 to 26 for equal-quarter-metres, as well as metres for with half-equal as also unequal quarters. The values of $p_i$ were fixed according to the general scheme presented in Table 2. By tradition, the note *pa* is considered as the middle note and therefore assigned the value 0.

We create an array p[] containing the $p_i$ values for each category of metres from 1 to 26 syllables per quarter and also for the half-equal and unequal metres. Consider the two examples depicted in Tables 3 and 4.

Table 3: Value of p[] for *Anuṣṭhup* metre (8 syllables per quarter)

| Quarters 1, 3 | 0 | 1 | 1 | 2 | 2 | 0 | 1 | 1 |
| --- | --- | --- | --- | --- | --- | --- | --- | --- |
| Quarters 2, 4 | 0 | 1 | -1 | 0 | 0 | 1 | 1 | 1 |

Table 4: Values of p[] for *Indravajrā, Upendravajrā, Upajāti* metres (11 syllables per quarter)

| Quarters 1, 3 | 0 | 0 | 1 | 2 | 2 | 0 | 0 | 1 | -1 | 0 | -1 |
| --- | --- | --- | --- | --- | --- | --- | --- | --- | --- | --- | --- |
| Quarters 2, 4 | 0 | 1 | 0 | 0 | 0 | 0 | -1 | 0 | 1 | 1 | 1 |

## 7. The Algorithm for Text-to-Tuneful-Speech

Initially, the verse is parsed by the metre classification algorithm and converted to its binary representation with *laghu* syllables taking the value 0 and *guru* syllables taking the value 1 [11]. For the sample verse given in Section 2 above, Table 5 depicts the binary representation thus obtained.

However, this may not match with the binary representation of individual units in the current context. Table 6 depicts the scenario generated by the consideration of individual units.

The reason for this discrepancy seen between Tables 5 and 6, is that in the current context depicted by Table 6, we are considering units as individual entities and independently categorizing them as *laghu* and *guru*. However, in the case of Table 5, we are considering each syllable in conjunction with the following one and making corrections to the *laghu-guru* status. For instance, "*van*" has a short vowel and is hence valued at 0 in Table 6. However, the full context of "*van*" is "*vande*", and so the double consonant "*nd*" after the vowel '*a*' forces the syllable "*van*" to be considered as *guru* and not *laghu* (as discussed in Section 5). Hence it is valued at 1 in Table 5.

Table 5: Actual binary representation of one quarter of the sample verse

| # | Syllable | Value (v) |
| --- | --- | --- |
| 1 | van | 1 |
| 2 | de | 1 |
| 3 | gu | 0 |
| 4 | rū | 1 |
| 5 | ṇāṁ | 1 |
| 6 | ca | 0 |
| 7 | ra | 0 |
| 8 | ṇā | 1 |
| 9 | ra | 0 |
| 10 | vin | 1 |
| 11 | de | 1 |

Table 6: Binary representation of one quarter of the sample verse when split into pronounceable units

| # | Syllable | Value (v) |
| --- | --- | --- |
| 1 | van | 0 |
| 2 | de | 1 |
| 3 | gu | 0 |
| 4 | rū | 1 |
| 5 | ṇāṁ | 1 |
| 6 | ca | 0 |
| 7 | ra | 0 |
| 8 | ṇā | 1 |
| 9 | ra | 0 |
| 10 | vin | 0 |
| 11 | de | 1 |

Now since we assign 1 unit of time to a *laghu* syllable and 2 units to a *guru* one, we adopt the following notation. Let $v_i$ be the actual value of unit $u_i$ (as per Table 5 and used for metre identification). We denote by $T_E$, the expected total units of time to chant the quarter of the verse under consideration, and by $T_A$, the actual total units of time to chant the concerned quarter. Therefore, we have

$T_E = \sum_{i=1}^{n}(v_i + 1)$

$T_A = \sum_{i=1}^{n}(t_i + 1)$

Clearly, $T_A \leq T_E$, because short syllables in the units may be converted to long if factors regarding the adjacent unit are taken into consideration. However, long syllables are never converted to short.

We present a general algorithm to adjust the beat of the chant, the chanting being achieved by concatenating the pre-recorded unit audio clips. One solution to making sure the beat is maintained, is to insert single units of silence when a unit is being intoned as *laghu* instead of *guru*. However, since a period of silence cannot be introduced in the middle of a word, in such cases, the syllabic unit being considered can be stretched to cover one more time unit.

**Algorithm** ConcatenateUnits
nTotalUnits = n;
if $T_A < T_E$ then
    k = 1;
    while k <= nTotalUnits do
        if $t_k < v_k$ then
            if k is the end of a word then
                insert 1 time unit of silence at position k+1;
            else
                stretch the kth audio unit;
            end if
            nTotalUnits = nTotalUnits + 1;
        end if
    end while
end if
n = nTotalUnits;
concatenate $f(u_1, p_1), f(u_2, p_2) \ldots f(u_n, p_n)$;
append 1 time unit of silence at caesura of the metre;
end **Algorithm**

## 8. The Overall Synthesis Algorithm

We now present the overall algorithm incorporating all factors discussed above.

**Algorithm** VerseTextToTunefulSpeechSynthesizer

**Step 1:** Parse the verse and identify its metre, caesura and retrieve the stored pitch array values;
**Step 2:** Apply *sandhi* rules to correct the specific cases where the pronunciation would change;
**Step 3:** Call Algorithm SplitVerseIntoUnits, to identify the pronounceable units in the verse;
**Step 4:** Retrieve the appropriate audio file for each unit from a file collection;
**Step 5:** Call Algorithm ConcatenateUnits, to adjust the beat of the chanting and also to apply the
note (pitch) variations to make the chanting tuneful;
**Step 6:** Play the concatenated file;
end **Algorithm**

The audio unit files were stored as .wav files and the concatenation was done by streaming them consecutively into a target .wav file. It was found that the function $f(u_i, p_i)$ used to change the frequency and hence the musical tone of the audio file may be realized both through the free APIs provided with the versatile open source software Audacity [15] and through the Microsoft DirectX SDK [17].

Clearly, the accuracy of the output is dependent on the tone and time duration of the recorded audio units. As such, during testing, changes to these audio unit files in terms of the intoning pitch and more importantly the time for which individual units were recorded, had to be made. This drastically improved the output quality.

## 9. Conclusions

Text-to-speech synthesis of Sanskrit verse is a hitherto unsolved problem. The fact is that such synthesis is problem-ridden owing to the numerous complexities inherent in the Sanskrit language in general and versification in particular. This work presents a beguilingly simple, yet comprehensive and effective solution based on the concatenative method of text-to-speech synthesis. The novel method presented here does not suffer from any performance bottlenecks.

This work utilizes earlier work by the authors on metrical classification of Sanskrit verses and on the Pāṇinian method of *sandhi* processing, and builds a Text-to-speech synthesizer for Sanskrit Verse. Empirical methods of analysis were used to create algorithms for splitting the verse into bits of pronounceable text and to significantly reduce the audio corpora required for the algorithm to function reasonably well. Furthermore, since Sanskrit verses are always tunefully chanted rather than uttered in a prosaic way, this solution incorporates a unique musical element too, and achieves a tuneful rendering of verses of various metres through manipulation of the frequency of the sound at appropriate places.

The work would be of tremendous use to those with visual impairments or reading disabilities, who would want to listen to and even memorize Sanskrit verses from any text they wish.

## References

[1] Alistair Conkie, Mark Beutnagel, Ann Syrdal, and Philip Brown, **Preselection of candidate units in a unit selection-based text-to-speech synthesis system**, Proceedings of ICSLP, volume 3, pages 314—317, 2000.

[2] Andrew Hunt and Alan Black, **Unit selection in a concatenative speech synthesis system**, Proceedings of ICASSP'96, volume 1, pages 373--376, Atlanta, GA, 1996.

[3] Anirban Lahiri, Satya Jyoti Chattopadhyay, Anupam Basu, **Sparsha: A Comprehensive Indian Language Toolset for the Blind**, Proceedings of the 7th international ACM SIGACCESS conference on Computers and accessibility, October 2005, (ISBN:1-59593-159-7), Pages: 114 – 120.

[4] Cyril Allauzen, Mehryar Mohri, and Michael Riley, **Statistical modeling for unit selection in speech synthesis**, Proceedings of the 42nd Annual Meeting on Association for Computational Linguistics, 2004, Article No.: 55

[5] Daniel Jurafsky, James H. Martin, **Speech and Language Processing**, Pearson Education 2000, Reprint 2005.

[6] Hideyuki Mizuno, Satoshi Takahashi, **Unit selection using k-nearest neighbor search for concatenative speech synthesis**, Proceedings of the 3rd International Universal Communication Symposium, 2009, Pages: 379-382 (ISBN:978-1-60558-641-0).

[7] Jonathan Allen, M. Sharon Hunnicutt, Dennis Klatt, **From Text to Speech: The MITalk system**, Cambridge University Press: 1987. (ISBN 0521306418).

[8] Mark Beutnagel, Mehryar Mohri, and Michael Riley, **Rapid unit selection from a large speech corpus for concatenative speech synthesis,** Proceedings of Eurospeech, volume 2, pages 607—610, 1999.

[9] Mattingly, Ignatius G., **Speech synthesis for phonetic and phonological models**, Thomas A. Sebeok (Ed.), Current Trends in Linguistics, Volume 12, Mouton, The Hague, pp. 2451-2487, 1974.

[10] Rama N., Meenakshi Lakshmanan, **A New Computational Schema for Euphonic Conjunctions in Sanskrit Processing**, IJCSI International Journal of Computer Science Issues, Vol. 5, 2009 (ISSN (Online): 1694-0784, ISSN (Print): 1694-0814), Pages 43-51, (http://ijcsi.org/papers/IJCSI-5-43-51.pdf - last accessed on 20.01.2010).

[11] Rama N., Meenakshi Lakshmanan, **A Computational Algorithm for Metrical Classification of Verse**, submitted to IJCSI International Journal of Computer Science Issues, Vol. 5, 2009 (ISSN (Online): 1694-0784, ISSN (Print): 1694-0814).

[12] Rubin P., Baer T., Mermelstein, P., **An articulatory synthesizer for perceptual research**, Journal of the Acoustical Society of America, 1981, 70, 321-328.

[13] Vaman Shivram Apte, **Practical Sanskrit-English Dictionary**, Motilal Banarsidass Publishers Pvt. Ltd., Delhi, 1998, Revised and Enlarged Edition, 2007.

[14] Van Santen P. H., Richard William Sproat, Joseph P. Olive, and Julia Hirschberg, **Progress in Speech Synthesis**, Springer: 1997. (ISBN 0387947019)

**Websites**

[15] Audacity Sound Editor and Recorder, **http://audacity.sourceforge.net** (last accessed on 17.01.2010).

[16] Indian Institute of Technology, Madras, **http://acharya.iitm.ac.in/disabilities/mbrola.php** (last accessed on 17.01.2010).

[17] Microsoft DirectX Developer Center, **http://msdn.microsoft.com/en-us/directx/default.aspx**, (last accessed on 17.01.2010).

[18] Technology Development for Indian Languages, Department of Information Technology, Ministry of Communication & Information Technology, Government of India, **http://tdil.mit.gov.in/standards.htm**, (last accessed on 17.01.2010).

**Dr. Rama N.** Completed B.Sc. (Mathematics), Master of Computer Applications and Ph.D. (Computer Science) from the University of Madras, India. She served in faculty positions at Anna Adarsh College, Chennai and as Head of the Department of Computer Science at Bharathi Women's College, Chennai, before moving on to Presidency College, Chennai, where she currently serves as Associate Professor. She has 20 years of teaching experience including 10 years of postgraduate (PG) teaching, and has guided 15 M.Phil. students. She has been the Chairperson of the Board of Studies in Computer Science for UG, and Member, Board of Studies in Computer Science for PG and Research at the University of Madras. Current research interests: Program Security. She is the Member of the Editorial cum Advisory Board of the Oriental Journal of Computer Science and Technology.

**Meenakshi Lakshmanan** Having completed B.Sc. (Mathematics), Master of Computer Applications at the University of Madras and M.Phil. (Computer Science), she is currently pursuing Ph.D. (Computer Science) at Mother Teresa Women's University, Kodaikanal, India. She is also pursuing Level 4 Sanskrit (*Samartha*) of the *Samskṛta Bhāṣā Pracāriṇī Sabhā*, Chittoor, India. Starting off her career as an executive at SRA Systems Pvt. Ltd., she switched to academics and currently heads the Department of Computer Science, Meenakshi College for Women, Chennai, India. She is a professional member of the ACM and IEEE.